\documentclass[11pt,letterpaper]{article}
 \pdfoutput=1
\usepackage[hyperref]{acl2017}
\usepackage{times}
\usepackage{hhline}
\usepackage{multirow}
\usepackage{graphicx}
\usepackage{color,soul}
\usepackage{xspace}
\usepackage{amsmath}
\usepackage{enumitem}
\usepackage{float}

\usepackage[labelformat=simple, font=small]{subcaption}

\usepackage[font=small]{caption}

\aclfinalcopy
\usepackage{hyperref}
\usepackage{xcolor}
\hypersetup{
    colorlinks = false,
    hidelinks = true
}

\usepackage[hang,flushmargin,perpage]{footmisc}

\newcommand{\crisis}{\textsc{crisis}\xspace}
\newcommand{\red}{\textsc{red}\xspace}
\newcommand{\amber}{\textsc{amber}\xspace}
\newcommand{\green}{\textsc{green}\xspace}
\newcommand{\urgent}{\textsc{urgent}\xspace}
\newcommand{\flagged}{\textsc{flagged}\xspace}
\newcommand{\liwc}{\textsc{liwc}\xspace}

\title{Triaging Content Severity in Online Mental Health Forums\Thanks{~~ This is a preprint of an article accepted for publication in Journal of the Association for
Information Science and Technology {\small\textcopyright} 2017 (Association for Information Science and Technology).}}
\author{
Arman Cohan$^{\spadesuit}$
~ {Sydney Young}$^{\spadesuit}$
~ {Andrew Yates}$^{\heartsuit}$
~ {Nazli Goharian}$^{\spadesuit}$\\
$^{\spadesuit}$Department of Computer Science, Georgetown University, Washington, DC, USA\\
$^{\heartsuit}$Max Planck Institute for Informatics, Saarland Informatics Campus, Saarbruecken, Germany\\
{\small \tt \{arman,nazli\}@ir.cs.georgetown.edu},\\
{\small \tt sey24@georgetown.edu, ayates@mpi-inf.mpg.de}
}

\begin{document}

\maketitle

\let\oldtabular\tabular
\renewcommand{\tabular}{\footnotesize\oldtabular}

\begin{abstract}
Mental health forums are online communities where people express their issues and seek help from moderators and other users. In such forums, there are often posts with severe content indicating that the user is in acute distress and there is a risk of attempted self-harm. Moderators need to respond to these severe posts in a timely manner to prevent potential self-harm. However, the large volume of daily posted content makes it difficult for the moderators to locate and respond to these critical posts. We present a framework for triaging user content into four severity categories which are defined based on indications of self-harm ideation.  Our models are based on a feature-rich classification framework which includes lexical, psycholinguistic, contextual and topic modeling features.
Our approaches improve the state of the art in triaging the content severity in mental health forums by large margins (up to 17\% improvement over the F-1 scores).
Using the proposed model, we analyze the mental state of users and we show that overall, long-term users of the forum demonstrate a decreased severity of risk over time. Our analysis on the interaction of the moderators with the users further indicates that without an automatic way to identify critical content, it is indeed challenging for the moderators to provide timely response to the users in need. 
\end{abstract}


\section{Introduction}

Mental health is an increasingly important health-related challenge in society; mental health conditions are associated with impaired health-related quality of life and social functioning \citep{saarni2007impact,strine2015depression}. Self-harm and suicide, as serious mental health conditions, are leading reasons of death world-wide \citep{nock2008cross,afsp2016suicide}. Each year an estimated number of 43,000 Americans die by suicide, on average there are 117 suicides per day, and about 500,000 people visit hospital for injuries due to self-harm \citep{karch2009surveillance,centers2015suicide,afsp2016suicide}.

Despite its pervasiveness, our understanding of suicide and self-harm related issues is limited. A notable reason for this is lack of large scale data on suicide. Most existing research on suicide is based on sparse curated data from a limited number of health-care centers. Furthermore, detecting and preventing potential self-harm acts remain a significant challenge due to reasons including lack of real-time data, privacy and confidentiality issues, and existence of bias in studies \citep{coppersmith2016exploratory}.

As social media usage has increased dramatically, individuals have tried to resolve their health problems by sharing them online, asking other users' opinions and seeking support. Therefore, social media have provided a valuable platform for large-scale analysis of mental health data and this analyses have offered great insights into mental health. Generally, it has been shown that social media can have broad applicability for public health research as the data from social media can reflect a variety of characteristics about individuals \citep{paul2011you,eichstaedt2015psychological}.

Online forums are a type of social media which are essentially communities in which users engage in discussion about topics of common interest. Mental health forums are centered around users who have directly or indirectly been involved in mental health conditions.

General social media platforms such as Twitter and Facebook are less topic-centric and more general purpose, in the sense that millions of users use them to discuss mundane events in their lives. While the signals coming from general social systems such as Twitter and Facebook are subtle and not directly about mental health, they are relevant and they have been previously utilized to support certain important tasks (e.g. \cite{coppersmith2015from,tsugawa2015recognizing,schwartz2014towards}). On the other hand, online forums are specifically designed for discussion around specific topics and they attract users with similar interests and goals \cite{de2014mental}. Some users in general social media such as Twitter can choose to be pseudonymous or anonymous, however, the identity of majority of the users are known. On the other hand, to protect their users, many online mental health forums such as ReachOut specifically ask their users to have anonymous profiles. The moderators in many of these forums further actively redact any post that could reveal the identity of the user. Such support for anonymity further encourages users to engage in sensitive mental health discussions and express their real thoughts and feelings. In this paper, we are focusing on online mental health forums as anonymous support platforms centered around people with similar experiences and problems.


There are three stages that lead to suicidal action among individuals who are in some sort of mental distress \citep{silverman1995prevention,Choudhury2016discovering}: 1- thinking, 2- ambivalence and 3- decision making. In the first two stages the individual is experiencing thoughts of distress, hopelessness, and low self-esteem. In the decision making stage, the individual might show explicit plans of taking their life. Individuals might seek support in any of these stages and online health forums are a ready platform enabling these individuals to ask for support. In many online mental health forums, there are moderators or more senior members who help the users with mental distress. Troubled users who are at risk of self-harm need to be attended to as quickly as possible to prevent a potential self-harm act. However, the volume of newly posted content each day makes it difficult for the moderators to locate and respond to more critical posts. Effective online manual triaging of all the forum contents is highly costly and not scalable.

We propose an approach for automated triaging of the severity of user content in online forums based on indication of self-harm thoughts. Triaging the content severity makes it possible for moderators to identify critical posts and help a troubled user in a timely manner to hopefully reduce the risk of self-harm to the user. We propose a feature-rich supervised classification framework that takes advantage of various types of features in the forums. The features include lexical, psycholinguistic, contextual, topic modeling, and dense representation features. We evaluate our approach on data provided by ReachOut\footnotemark[1], a large mental health forum. We show that our approach can effectively identify the critical content which will assist the moderators in attending to the in-need users in a timely manner. We show that without an automatic way for identifying critical posts, the moderator's response time does not correlate with the severity of the posts, which further confirms that manually identifying these posts is a challenge for moderators. Finally, analysis of the user content on this forum shows that on average, the content severity of users tends to decline as they interact with the forum which is evidenced by the transition from more critical to less critical content.

\footnotetext[1]{www.ReachOut.com}

The contributions of this work are as follows: \textit{(i)} an effective approach for triaging the content severity in online mental health forums based on indication of self-harm ideation; \textit{(ii)} providing insight into the effect of online mental health forums on users through analysis of their content; \textit{(iii)} analyzing the interaction of moderators with users; and \textit{(iv)} extensive evaluation of the proposed approach on a real-world dataset.


\section{Related Work}

\subsection{Healthcare and Mental Health through Social Media}
In recent years, healthcare has benefited enormously from social media data \citep{dredze2012social}. Many studies have investigated public health surveillance by utilizing the Twitter public data \citep{paul2011you,lamb2013separating,Parker2015,Chen2016,paul2016social}. Results of these studies show consistency with other information resources for public health such as official reports released by governments, reports released by Centers for Disease Control and Prevention (CDC) and other online sources such as Google Flu Trends\footnotemark.

\footnotetext{\url{https://www.google.org/flutrends/}}

Social media has also become a popular platform for people with mental health conditions to express their feelings and seek support from other users. It has helped individuals with depression by providing them means to connect to people with shared experiences who can answer their questions \citep{dao2015nonparametric,olteanu2016towards}. Consequently, the information from social media has become a significant resource providing more insight into psychological and mental conditions and problems. \citet{deChoudhury2013major} explored social media to identify and diagnose depression among individuals. They analyzed the posting of a set of Twitter users through time and identified signals for characterizing the onset of depression in individuals. \citet{park2013perception} showed that depressed individuals perceived social media (Twitter) as a tool for social awareness and emotional interaction while non-depressed individuals are mostly regular information consumers. \citet{schwartz2014towards} used Facebook data to build a regression model to predict degree of depression in individuals.   \citet{portier2013understanding} conducted sentiment analysis on the cancer survivor forum content and compared the sentiment change of the user content before and after interaction with the community.  There exist many other works on analysis of social media for mental health problems such as depressive disorders \cite{deChoudhury2013social,tsugawa2015recognizing}, addiction \citep{murnane2014unraveling}, insomnia \citep{jamison2012can}, schizophrenia \citep{mitchell2015quantifying} and various other conditions \citep{coppersmith2015from}.

While many of the aforementioned mental health disorders are closely related to suicidal behaviors and could lead to suicidal ideation, our focus in this paper is to identify the severity of the content based on indication of self-harm risk to individuals.

\subsection{Social Media and Suicide}
Previous work has studied self-harm and suicidal behavior through text analysis. Some researchers explored the language usage in content relating to suicide to identify signals of this behavior to predict suicidal actions. \citet{thompson2014Predicting} predicted the risk of suicide in military personnel and veterans using the clinical notes and online social media data (Facebook posts). They used a model based on Random Forest classifier \citep{breiman2001random} with bag-of-words features. \citet{jones2007development} developed statistical prediction rules to discriminate between genuine and simulated suicide notes. \citet{lester2010final} analyzed the language of suicide notes to better understand suicidal behaviors in individuals.
\citet{coppersmith-EtAl:2016:CLPsych} examined data from Twitter users who have attempted to take their life and provided an exploratory analysis of patterns in language around their attempt.  .
Some researchers have analyzed suicidal behaviors through detecting sentiment and emotional variations of the content \citep{cherry2012binary,pestian2012sentiment,desmet2013emotion}.
Prior work has also explored classification of suicidal content. \citet{burnap2015machine} proposed an ensemble classification approach to classify tweets into suicide related topics such as suicidal ideation, reporting of a suicide, memorial, campaigning and support. \citet{braithwaite2016validating} conducted a user study on a group of individuals and analyzed their Twitter posts using Decision Tree classifier to differentiate individuals with higher suicide risks from individuals who are not at risk. Finally \citet{Choudhury2016discovering} proposed that social media could be used to predict shifts from mental health discussions to expression of suicide thoughts. Specifically, they analyzed language in Reddit\footnotemark mental health community and employed a framework based on propensity score matching \citep{rosenbaum1984reducing} to predict suicidal shifts in users. Unlike these works, our focus is triaging the content severity in mental health online forums based on the risk of self-harm to the users.

\footnotetext{https://www.reddit.com/}

The closest work to ours is the recent shared task \citep{milne2016overview} on automatic identification of content severity in mental health forums by the 2016 Computational Linguistics and Clinical Psychology Workshop \citep{clpsych-workshop-2016}. 16 teams participated in this challenge and a variety of methods have been proposed. Most of the systems, generally used Support Vector Machine (SVM) classifiers \citep{Cortes1995} or an ensemble of some other standard classifiers for identifying the content severity. We briefly describe the top 3 approaches: \citet{kim-EtAl:2016:CLPsych} used a Stochastic Gradient Decent classification framework. They utilized the body of the text as the main source for feature extraction and represented the post by weighted TF-IDF\footnotemark unigrams and distributed representation of documents \citep{le2014distributed}. \citet{malmasi-zampieri-dras:2016:CLPsych} used a hierarchical classification framework. They employed a Random Forest meta-classification approach on top of a set of base classifiers. Finally, \citet{brew:2016:CLPsych} used SVM with Radial Basis Function (RBF) kernel; they utilized TF-IDF unigram and bigram features, author type, post information and position of the post in the thread as the features for the classifier.

\footnotetext{Term Frequency - Inverse Document Frequency}

In contrast to these works, our approach is feature-rich; many features that we use are not present in the aforementioned prior work, such as psycholinguistic, contextual, topic modeling and skip thought features (see the Methods section for details). We also utilize an ensemble classifier using different subsets of features. Our proposed model outperforms the state-of-the-art by large margins.

This work extends our earlier effort in the CLPsych workshop where we used a Logistic Regression classifier to identify the severity of the posts \citep{cohan-young-goharian:2016:CLPsych}. We achieve up to 24\% F1 score improvements over our previous results at CLPsych 2016. The improvements are due to utilizing a better learning algorithm, extending the feature sets and introducing our new ensemble model. Our models outperforms the state-of-the-art by large margins.

While the aforementioned works only focus on triaging the content severity, we further utilize the triaging model to perform analysis of user interactions in this forum to gain insight on the impact of the forum on the users with mental health issues. We analyze the moderators' response time to users and show that without an accurate and efficient content triaging system, manually identifying severe posts in forums with large number of users is indeed difficult.


\section{Severity Triaging}
\label{sec:triaging-def}

Our main objective is to determine the severity of the mental health forum posts based on signs of self-harm thoughts in the content. Triaging content severity enables moderators to attend to severe cases in a timely manner and hopefully prevent a potential self-harm attempt.

Our approach for triaging the content severity is a supervised learning framework. In the following, we first define the severity categories, then we explain the features that we use for the classification and finally, we describe the learning algorithm.

\subsection{Severity categories}

We consider the following 4 levels of severity for the post content, as defined by \citep{milne2016overview}:
\begin{itemize}
\item{\textbf{Green -}} posts that do not show any signs or discussions about self-harm and thus do not require direct input from the moderators. These posts are usually general statements or follow up discussions that do not reflect any major concern.
\item{\textbf{Amber -}} posts that include minor clues that might indicate signs of struggle by the user. These posts need the moderator's attention at some point, but prompt intervention is not necessary.
\item{\textbf{Red - }} posts indicating that the user is in acute distress and moderators should attend to them as soon as possible.
\item{\textbf{Crisis - }} posts indicating that the user is in imminent risk of self-harm. These posts could be about the authors themselves or someone that the author of the post knows. Moderators should prioritize these cases above all others.
\end{itemize}

Table \ref{tab:example-cat} shows synthesized examples of posts in each of these severity categories\footnote{The provided examples throughout this paper are very similar to the ones in the ReachOut forum. According to the data collection policies on protecting users' identities, we are unable to include the exact posts from the forum.}. Following the terminology used by \citet{milne2016overview}, we consider the union of \crisis, \red and \amber categories as \flagged posts, because they indicate that user might be at risk and needs attention at some point. Similarly, we consider the union of two more critical categories, i.e \crisis and \red as \urgent.

\begin{table*}[]
\centering
\begin{tabular}{|c|c|c|c|}
\hline
\green                                                                                                                         & \amber                                                                                                                                                & \red                                                                                                            & \crisis                                                                                                                               \\ \hline
\begin{tabular}[c]{@{}c@{}}I'm proud that I was \\ able to call and\\ keep up a phone\\ conversation with\\ my mum.\end{tabular} & \begin{tabular}[c]{@{}c@{}}There are so many stuff\\ I'm thinking about, but my\\ medications are slowing\\ my thoughts down\\ and making it\\ more manageable\end{tabular} & \begin{tabular}[c]{@{}c@{}}I feel helpless and\\ things seem\\ pointless. I hate\\ feeling so down…\end{tabular} & \begin{tabular}[c]{@{}c@{}}I’m having some\\ strong thoughts about\\ ending my life,\\ nothing helps.\end{tabular} \\ \hline
\end{tabular}
\caption{Example of posts in each severity category.}
\label{tab:example-cat}
\end{table*}

Due to large volume of posts produced each day, it is not possible for moderators to identify all the critical posts in a timely manner. Our goal is to predict the severity of the forum posts' content so that the moderators can locate critical cases and attend to them as soon as possible.
We propose a feature-rich machine learning approach utilizing psycholinguistic, topic modeling and contextual features.

\subsection{Features}

Since the forum posts are written in unstructured raw text, we extract representative features from the text that are helpful for the supervised learning. Particularly, we extract the following categories of features:

\begin{itemize}[wide, labelwidth=!,labelindent=0pt]

\item \textbf{Bag of words~~} An standard approach for text representation is to model the text with bag of its constituent words. This results in a sparse vector for each text in which each element associates with a word in the vocabulary and is weighted according to some weighting scheme. We use the unigram and bigram bag of words representation of text with frequency of terms as their weights. Throughout the paper, when we refer to some textual content (e.g. post body) as features, we are essentially referring to the unigram and bigram bag of words representation of that text, unless otherwise noted. Before representing the text with bag of words features, we perform standard minimal preprocessing on it by lowercasing and removing stopwords.

\item \textbf{Psycholintuistic~~} The psycholinguistic features are meant to capture the different dimensions of a user's mental state through analysis of their language usage.

\begin{itemize}[wide, labelwidth=!,leftmargin=0pt,labelindent=2pt]

\item \textit{LIWC}: Linguistic Inquiry and Word Count (LIWC) \citep{pennebaker2015development} is a tool that captures quantitative data regarding various psychological dimensions given the user's textual writings. It utilizes several psychological lexicons along with a text analysis module that associates text with different psychologically-relevant categories. We use this tool to extract different psychological attributes from the language expressed in the users' posts. While LIWC provides over 100 distinct attributes, our experimentation showed that the affective attributes, drive attributes, tonality, informal language usage, anxiety attributes and negation are the most helpful for this task.

\item \textit{Emotions}: Emotions are very closely related to suicide. Therefore, the emotion that is reflected by the post can be a good indicator about level of severity of the content. For example, if a user's post indicates the ``anger'' emotion, it is more likely to be severe in comparison with a post that shows the ``happiness'' emotion. To quantify the emotions associated with a specific post, we use DepecheMood \citep{staiano2014}, a lexicon with emotional probabilities associated with more than 37000 terms. The emotions considered by the lexicon are ``fear'', ``amusement'', ``anger'', ``annoy'', ``apathy'', ``happiness'', ``inspiration'' and ``sadness''. To obtain the overall distribution of emotion over these categories for a post, we average the emotion distribution of all words in the post to obtain probability of each emotion given the post. We use these probabilities as features for the classification. In addition to the specific probabilities, we also consider the dominant emotion of the post as a separate feature.

\item \textit{Subjectivity}: Similarly, subjective posts are more likely to be related to a severe post than an objective post. We utilize the MPQA subjectivity lexicon \citep{wilson2005recognizing} to differentiate between the subjective and objective posts. This lexicon contains contextual subjectivity about words or phrases that indicates expression of an emotion, opinion, stance, etc.

\end{itemize}

\item \textbf{Contextual~~} One characteristic of online forums is that they are designed to support user discussion. Therefore, having information about the context of a given post in the discussion thread provides additional information about its content. We extract the following contextual features:

\begin{itemize}[wide, labelwidth=!,leftmargin=0pt,labelindent=2pt]

\item \textit{Author's prior posts}: Author's prior posts in the thread captures the development of thoughts by the user and also in combination with the body of the post captures whether the post deviates from the author's prior posts in a significant way.

\item \textit{Prior discussion}: The posts preceding a target post and written by other users help in capturing surrounding discussion and development of thoughts for the target user. Specifically, we consider a window of 3 posts by other users preceding the target post as the context of the post in the thread. Limiting the window size to 3 is due to our observation that in long threads, the discussion usually deviates after a few posts, hence considering all the posts would introduce noise to the model\footnotemark. We could also consider the posts succeeding the target post as additional features, however, that would not correspond to a real-world scenario. In a realistic setting, the goal is to triage the content on the forum as soon as they are posted and therefore, to comply with this setting, we do not consider any features relating to content submitted after the target post.

\footnotetext{We experimented with context window of sizes 1 to 5. The best performance was for context size of 3, therefore we chose window of 3 posts as the context size.}

\item \textit{Last sentence}: Finally, some critical posts are long, and mostly about some mundane and usual events that happen; in these posts, there is a sudden change at the end of the post indicating that the user might be at risk. Take the following example which is a snippet from the beginning and ending part of a longer post (Parts indicated with [...] are omitted for brevity):

\textit{
``Now, I think we all know what it's like to be rejected by friends, dates, etc. While I have been stood up by a certain friend a few times, this really got to me. My dad said on tuesday [...] \\
... I woke up today and I since morning just don't know what to do anymore. I feel like I have nothing to live for and nothing makes me happy anymore.''}

In this example, most of the body of the post does not indicate any immediate risk to the user. However, this sudden change in the user's mental state shows that this content is potentially a severe case. If we only rely on the features capturing the entire post, the mental state shift will not be apparent as most of the post do not show any signs of risk. Therefore, we also consider the last sentence as a separate feature; we utilize the \liwc attributes for the last sentence to focus on the final mental state of the user and to eliminate some of the dilution that may occur in longer posts.

\end{itemize}

\item \textbf{Topic modeling~~} We use the abstract ``topics'' that occur in the collections of posts as another set of features for classification. Topic modeling \cite{blei2012probabilistic} is a widely used approach for discovering the latent semantic structures (``topics'') in a text body. Latent Dirichlet Allocation (LDA) \cite{blei2003latent} is a generative model that describes how the documents in a dataset are created. A brief description of the LDA generative process is as follows:

\begin{enumerate}
\item For each document:
\begin{enumerate}
\item\label{lda:step} Draw a distribution over topics
\item Generate each word in the document by:
\begin{enumerate}
\item Drawing a topic $\beta_j$ according to the distribution selected in step (a).
\item Drawing one word from the $V$ words in the topic $\beta_j$
\end{enumerate}
\end{enumerate}
\end{enumerate}

Using this generative process, the LDA model tries to find a set of topics that are likely to have generated the collection. We trained the LDA topic model on the entire forum posts to obtain the latent topics associated with each post and we used these topics as additional features\footnotemark.

\footnotetext{We limited the number of topics to 100. We experimented with 20,50,100, and 200 topics and 100 topics was the optimal choice.}

\item \textbf{Skip thought vectors~~} Bag of words representation of the post is a sparse representation in which most of the entries are zero. More recently, approaches have been proposed for obtaining a dense representation of sentences that can encode syntactic and semantic properties of sentences in vectors. Skip thought vectors \citep{kiros2015skip} are one such model that use ``sequence to sequence'' models on pairs of consecutive sentences to learn the sentence encoding. Their model consist of a encoder-decoder framework in which the encoder maps words to a sentence vector and a decoder is used to generate the surrounding sentences. By analysis through several tasks, \citet{kiros2015skip} showed that this approach results in good sentence encodings when trained on a sufficiently large corpus. We use this model to encode the forum posts in dense representations. We average the vector representation of all sentences in the post to encode the entire post.

\item \textbf{Forum metadata ~~}
Forum metadata such as number of post views, length of the thread, and number of post ``kudos'', a ReachOut feature similar to ``likes'' on Facebook, are additional features that we considered. Motivated by previous research that identified the time of day of online activity as a useful mental health signal \citep{coppersmith2014quantifying,de2013predicting}, we also consider the broad temporal categories (day and night) as well as more fine-grained intervals (morning, afternoon, evening, and night). However, we did not observe an increase in the classifier's performance with the addition of the temporal metadata attributes.


\end{itemize}

\subsection{Learning algorithm}

After extracting features, we use supervised multi-class classification for triaging the user posts into different severity categories. We use the XGBoost Tree Boosting \citep{chen2016treeboosting} as the learning algorithm. We experimented with several other standard classifiers such as logistic regression, random forest, and SVM, but XGBoost showed the best results.

Let the dataset $D=\{\mathbf{x}_i,y_i\}_{i=1}^n$ consist of $n$ different training instances in which the $i\;$th instance is represented by a feature vector $\mathbf{x}_i$ and label $y_i$. In matrix notation, the entire feature vector and the labels are represented as $(\mathbf{X},\mathbf{y})$. Given this dataset $D$, the XGBoost tree ensemble model uses an ensemble of $K$ additive functions (regression trees) to predict the output $\hat{y}_i$:

\begin{equation}
\hat{y}_i = \phi(\mathbf{x}_i) = \sum\limits_{k=1}^K f_k(\mathbf{x}_i), f_k \in \mathcal{F}
\end{equation}

where $\phi$ represents the model that predicts the output given the feature vector $\mathbf{x}_i$, $\mathcal{F}$ is the space of all regression trees, and $K$ is the total number of regression trees used. The essential part of the model is regression trees $f_i$. To learn $f$, given the model output $\hat{\mathbf{y}}$ and the true class labels $\mathbf{y}$, the following regularized objective function is optimized over the training data:

\begin{equation}
\label{eq:2}
\mathcal{L} = \sum\limits_{i=1}^{n} l(y_i, \hat{y}_i) + \sum\limits_{k=1}^{K}\Omega(f_k)
\end{equation}

where $l$ is a differentiable convex loss function (e.g. squared loss $l(y_i, \hat{y}_i)=(y_i - \hat{y}_i)^2$), and $\Omega(f_k)$ is the regularizing function that penalizes the complexity of the functions to prevent overfitting. The model is trained additively by greedily adding $f_k$ that most improves the model based on equation \ref{eq:2}. The additive function $f_k$ is also learned by a greedy tree growth algorithm. Several approximations are used that can quickly optimize the objective function. For more details on these steps, refer to the XGBoost reference \cite{chen2016treeboosting}.

 In addition to the single classification model, we also utilize the ensemble of several XGBoost classifiers, each trained on a different subset of features from the entire feature space. We empirically determine the optimal subsets of features. By ensembling, we use multiple classifiers to obtain better performance than individual classifiers. Intuitively, we take advantage of several conceptually different models (each of which obtained by training on a different feature set), and we aggregate their predictions to obtain the final class label. We use the majority voting ensembling approach which has been shown to balance out the weaknesses of individual classifiers \cite{lam1997application,opitz1999popular}.

 \begin{table*}[t]
     \small
 \centering
 \begin{tabular}{|c|cc|cc|cc|}
 \hline
                   & \multicolumn{2}{c|}{Train set} & \multicolumn{2}{c|}{Test set} & \multicolumn{2}{c|}{Total} \\ \hline
 Severity Category & \# posts       & \% posts      & \# posts      & \% posts      & \# posts     & \% posts    \\ \hline
 \crisis            & 39             & 4             & 1             & 0             & 40           & 3           \\
 \red              & 110            & 12            & 27            & 11            & 137          & 12          \\
 \amber            & 249            & 26            & 47            & 19            & 296          & 25          \\
 \green            & 549            & 58            & 166           & 69            & 715          & 60          \\  \hline
 Total             & 947            & 100           & 241           & 100           & 1188         & 100         \\ \hline
 \end{tabular}
 \caption{Distribution of the labeled forum posts in the dataset. Percentages are rounded.}
 \label{tab:data-characteristics}
 \end{table*}

Formally, let \{$\phi^{(1)}$, ..., $\phi^{(m)}$\} be $m$ models obtained by training the classifier on $m$ different feature sets \{$\mathbf{X}^{(1)}$, ... ,$\mathbf{X}^{(m)}$\}. Similarly let $\{\hat{\mathbf{y}}^{(1)}$, ..., $\hat{\mathbf{y}}^{(m)}\}$ represent the output predicted by models \{$\phi^{(1)}$, ..., $\phi^{(m)}$\}. For the $i\;$th instance in the dataset, the majority voting ensembling approach predicts the class label $\hat{y}_i$ according to the following:

 \begin{equation}
    \hat{y}_i = \underset{c\in\{c_1,...,c_T\}}{\mathrm{argmax}}\big(\Big|\{j\in\{1,...,m\}:\hat{y}_i^{(j)}=c\}\Big|\big)
 \end{equation}

where $\{c_1,...,c_T\}$ is the set of all possible class labels.

XGBoost has several hyperparameters including the learning rate ($\eta$), the minimum sum of the weigths of all observations in a child ($min$-$weight$), and the maximum depth of the tree ($max$-$depth$). We used the default parameters which are $\eta=0.3$, $min$-$weight=1$ and $max$-$depth=6$. We did not observe any performance gain by modifying the default recommended  hyperparameters.


\section{Experimental setup}
\label{sec:data}

\subsection{Data}
The data that we use in this research are forum posts from ReachOut.com which is a very large and popular mental health forum in Australia and receives about 1.8 yearly visits \citep{millen2015annual}. While this forum provides a discussion platform for ordinary topics such as life, family and friendship, its main purpose is to support discussions around more critical topics such as addiction, sexuality, identity and mental health problems. Most of the users and visitors are young people aging between 14 to 25 years old. ReachOut employs several senior moderators as well as younger people who volunteer for forum moderation. These moderators focus on cases that require attention and try to help these individuals by engaging in the discussion, showing compassion and support, and providing links and resources to the individuals.

We use a subset of the ReachOut forum containing 65,755 posts, 1,188 of which had been labeled by moderators based on 4 different categories of severity. The dataset contains separate training and testing sets; its characteristics are outlined in Table \ref{tab:data-characteristics}. The posts occurred between July 2012 and June 2015, with labeled posts being from May 2015 to June 2015. The posts were written by 1,647 unique authors. Each post contains several fields such as the post date and time, username of the author, number of kudos, subject of the thread, and the textual body of the post.

\paragraph{Data collection.}  The full details of the data collection and the discussion on the ethical issues are discussed by \citet{milne2016overview}. While analysis of the mental health forum data provides many benefits, there are always trade-offs between the benefits and the risk to the privacy of the individuals. \citet{milne2016overview} identified three groups of participants to whom the data collection and annotation process could cause harm: to the researchers who annotated the data, to the researchers who accessed the data, and to the people who authored the content. The data collection process ensured that the researchers were aware of the distressing nature of the content. To protect its users, forum members of the ReachOut are instructed to keep themselves safe and anonymous. Furthermore, the moderators in the forum actively redact any content that might reveal the identity of the users. The organizers further protected the forum member's anonymity by restricting researchers in contacting the individuals in the forum, distributing the data, and cross-referencing individuals against other social media.



\begin{table*}[t]
  \small
    \begin{subtable}{1\linewidth}
    \centering
        \begin{tabular}{|l|cc|cc|cc|}
        \hline
        \multirow{2}{*}{Methods}                          & \multicolumn{2}{c|}{\begin{tabular}[c]{@{}c@{}}Macro Average over\\ non-\green categories\end{tabular}} & \multicolumn{2}{c|}{\begin{tabular}[c]{@{}c@{}}\flagged\\ vs. \green\end{tabular}} & \multicolumn{2}{c|}{\begin{tabular}[c]{@{}c@{}}\urgent\\ vs non-\urgent\end{tabular}} \\ \cline{2-7}
         & F1              & Acc             & F1            & Acc          & F1           & Acc          \\ \hline
        Baseline                    & 31              & 78              & 75            & 86           & 38           & 89           \\
        \citet{cohan-young-goharian:2016:CLPsych} &  41 & 80 & 81 & 87 & 67 & 92 \\
        \citet{brew:2016:CLPsych}                 & 42              & 79              & 78            & 85           & 69           & 93           \\
        \citet{malmasi-zampieri-dras:2016:CLPsych}        & 42              & 83              & 87            & 91           & 64           & 93           \\
        \citet{kim-EtAl:2016:CLPsych}            & 42              & 85              & 85            & 91           & 62           & 91           \\ \hline
        This work (Single model) & 47.2            & 93.9            & 90.0          & 91.7 & 73.1         & 92.9         \\
        This work (Ensemble model)   & \textbf{50.5}   & \textbf{94.7}   & \textbf{92.2} & \textbf{93.4}         & \textbf{75.5} & \textbf{94.6}         \\ \hline
        \end{tabular}
    \caption{}
    \label{tab:classification-res-test}
    \end{subtable}
\bigskip
    \begin{subtable}{1\linewidth}
      \setlength{\tabcolsep}{2pt}
    \centering
    \begin{tabular}{|l|c|c|c|c|c|c|}
    \hline
    \multirow{2}{*}{Methods}                                              & \multicolumn{2}{c|}{\begin{tabular}[c]{@{}c@{}}Macro Average over\\ non-\green categories\end{tabular}} & \multicolumn{2}{c|}{\begin{tabular}[c]{@{}c@{}}\flagged\\ vs. \green\end{tabular}} & \multicolumn{2}{c|}{\begin{tabular}[c]{@{}c@{}}\urgent \\ vs. non-\urgent\end{tabular}} \\ \cline{2-7}
                                                                          & F1                                                 & Acc                                                & F1                                       & Acc                                     & F1                                          & Acc                                       \\ \hline
    Baseline                                                              & 29.0                                     & 87.4                                     & 78.2                           & 80.6                          & 64.2                           & 86.7                            \\ \hline
    This work (single model)   & 43.0 $\dagger$                                    & 89.6 $\dagger$                                   & 85.1 $\dagger$                          & 86.1 $\dagger$                         & \textbf{78.3} $\dagger$                   & 90.8 $\dagger$                           \\ \hline
    This work (ensemble model) & \textbf{44.5}  $\ddagger$                           & \textbf{90.6}  $\ddagger$                           & \textbf{88.1}  $\ddagger$                 & \textbf{88.8}  $\ddagger$                & 77.6 $\dagger$                             & \textbf{91.4} $\dagger$                  \\ \hline
    \end{tabular}
    \caption{}
    \label{tab:classification-res-cv}
\end{subtable}
\caption{Results of triaging content severity. Numbers are percentages. \flagged category is \amber $\cup$ \red $\cup$ \crisis. \urgent category is \red $\cup$ \crisis. F1 is F1-Score and Acc is Accuracy. Baseline is the SVM classifier on post body (unigram and bigram features). Table (a) presents classification results and comparison with the baseline and state-of-the-art on the test set. Table (b) shows classification results on training set based on 10-fold stratified cross validation. For Table (b), $\dagger$($\ddagger$) shows statistically significant improvement over the baseline (all other methods in the Table) according to the Student's t-test ($p<0.02$).}
\label{tab:classification-res}
\end{table*}


\subsection{Evaluation}
Following \citet{milne2016overview}, we use the accuracy and F-1 scores for evaluating the classification performance to be able to directly compare the performance of our approach with the state-of-the-art. To aggregate the scores for the individual categories, \citet{milne2016overview} used the macro average of F-scores for the non-\green (critical) categories as the official metric for the CLPsych 2016 shared task. This metric emphasizes the importance of triaging among the critical categories. They also consider the F-1 and accuracy scores for binary classification of \flagged (i.e. \crisis $\cup$ \red $\cup$ \amber) vs. \green, and \urgent (i.e. \crisis $\cup$ \red) vs. non-\urgent categories to capture the performance of systems in identifying critical posts. We also use these additional metrics to further evaluate the performance of our approach. \flagged classification shows that the post contains content indicating risk of self-harm to the user while \urgent indicates that the user is at a more imminent risk and needs prompt attention (see the Method Section for complete definitions of severity categories).


\begin{table*}[t]
\centering
\begin{tabular}{|l|l|}
\hline
\multicolumn{1}{|c|}{Model} & \multicolumn{1}{c|}{Features}                                                                                                                                                                                                                                                                                                                                                        \\ \hline
Single model            & \begin{tabular}[c]{@{}l@{}}Post body, forum metadata, subjectivity, emotions,\\ contextual features, last sentence, topic modeling, LIWC\end{tabular}                                                                                                                                                                                                                                 \\ \hline
Ensemble model              & \begin{tabular}[c]{@{}l@{}}1- Post body, forum metadata, subjectivity, emotion\\ 2- Post body, contextual features, emotion features, LIWC\\ 3- Post body, contextual features, last sentence\\ 4- Post body, last sentence, emotion, sentiment\\ 5- Post body, contextual features, topic modeling\\ 6- Post body, contextual features, LIWC, clue words, forum metadata\end{tabular} \\ \hline
\end{tabular}
\caption{Features in our single and ensemble models. The ensemble model is comprised of 6 classifiers with fewer number of features.}
\label{tab:features}
\end{table*}


\subsection{Baselines and comparison}
We compare our methods with the top 4 performing systems among 16 total participating teams in the CLPsych 2016 shared task. To better evaluate our methods, we also consider a simple baseline which is SVM classifier with unigram and bigram bag-of-words features extracted from the body of the post (refer to bag-of-words features explained in Methods section for details).


\section{Results and analysis}

The results of our models for triaging the content severity compared with the baseline and state of the art systems is presented in Table \ref{tab:classification-res}; it includes results on the test set \ref{tab:classification-res-test}, as well as stratified cross-validation\footnote{The stratified cross validation in contrast to the regular cross validation preserves the distribution of the classes when splitting the data into train and test sets.} results on the training set \ref{tab:classification-res-cv}. For prior work, we report the official results that are percentages without any precision points. The single model indicates the performance of our proposed model using a single classifier while the ensemble model is a model based on 6 different classifiers. The features used in each of the models are presented in Table \ref{tab:features}. In the Analysis Section, we will discuss the effect of different features on the performance. As illustrated in Table \ref{tab:classification-res-test}, our models outperform the baseline and all top performing state of the art systems by large margins. We observe that the non-\green macro average F1 score for the individual and ensemble models improves over the best system \citep{kim-EtAl:2016:CLPsych} by +12\% and +17\%, respectively. Similarly, we observe that the F1 scores for the \flagged category is 3\% and 5\% higher than the best system with the individual and ensemble models, respectively. Finally, in \urgent category, the individual and ensemble models achieve 73.1\% and 75.1\% F1 scores respectively, which shows large improvement over the state of the art. We observe similar improvements in the cross-validation results on the training set (Table \ref{tab:classification-res-cv}). Since we have 10 different folds on the training set, we also perform a statistical significance test and we observe statistically significant improvement over the baseline for both the single and ensemble methods (Student's t-test); the ensemble method also outperforms the single method statistically in virtually all metrics. In particular, the single and ensemble models achieve 48\% and 53\% improvements over the baseline based on non-\green macro average F1 scores.


\begin{table*}
  \small
\centering
\begin{subtable}{1\linewidth}
    \centering
    \begin{tabular}{|l|c|c|c|c|}
    \hline
                                                    & \multicolumn{4}{c|}{Severity categories}                  \\ \cline{2-5}
    Methods                                         & \crisis (1) & \red (27)     & \amber (47)   & \green (69) \\ \hline
    Baseline                                        & 0           & 39            & 53            & 90          \\
    \citet{cohan-young-goharian:2016:CLPsych} & 0 & 59 & 64 & 90 \\
    \citet{brew:2016:CLPsych}                  & 0           & 65            & 61            & 88          \\
    \citet{malmasi-zampieri-dras:2016:CLPsych} & 0           & 58            & 69            & 93          \\
    \citet{kim-EtAl:2016:CLPsych}              & 0           & 65            & 61            & 94          \\ \hline
    Single model                          & 0           & 67.7          & 67.4          & 93.7        \\
    Ensemble model                            & 0           & \textbf{75.5} & \textbf{76.1} & \textbf{95.2}  \\ \hline
    \end{tabular}
    \caption{}
    \label{tab:fingrained-results-test}
\end{subtable}
\begin{subtable}{1\linewidth}
  \centering
  \begin{tabular}{|l|c|c|c|c|}
\hline
Methods                                                               & \crisis                    & \red                      & \amber                    & \green                    \\ \hline
Baseline                                                              & 5.3             & 31.5           & 50.7            & 85.5            \\ \hline
This work (single model)   & 17.0$\dagger$            & 53.0 $\dagger$          & 63.2 $\dagger$           & 89.0 $\dagger$           \\ \hline
This work (ensemble model) & \textbf{21.3} $\ddagger$  & \textbf{55.3} $\ddagger$  & \textbf{69.1} $\ddagger$  & \textbf{91.1} $\dagger$  \\ \hline
\end{tabular}
  \caption{}
  \label{tab:finegrained-results-train}
\end{subtable}
\caption{Fine-grained classification results for each severity category. Numbers show macro-average F-1 scores in percentages. Last two rows show models proposed in this work. The top table (a) shows classification results and comparison with the baseline and state of the art based on each severity category on the test set. The numbers in parenthesis in front of each category is the total number of instances in that category. Note that \crisis has only 1 instance and no system was able to detect that. Table (b) shows classification results by severity category on the training set (10-fold stratified cross validation). For Table (b), $\dagger$($\ddagger$) shows statistically significant improvement over the baseline (all other methods in the Table) according to the Student's t-test ($p<0.02$).}
\label{tab:per-category}
\end{table*}


Table \ref{tab:per-category} shows the breakdown of results by each category. We present results on the test set in Table \ref{tab:fingrained-results-test} and cross validation results on training set in Table \ref{tab:finegrained-results-train}. It should be noted that there was only 1 \crisis case in the test set and no team out of 16 teams were able to correctly identify this case. While our models were also unable to find the single \crisis case, they show improvements over the state of the art in other categories. Specifically, we observe that the ensemble model achieves F1 score of 75.5\% in \red which improves over the best performance (65\%) by 16\%. Similarly, we observe large improvement of F1 for the \amber category (10\%). Finally, our model also slightly improves upon the state of the art on the \green category. We also report results on the training set evaluated by 10 fold stratified cross validation (Table \ref{tab:finegrained-results-train}). As illustrated, our methods achieve statistically significant improvement over the baseline in all severity categories. The overall lower performance on the \crisis category is mainly due to the limited training data in this category. As shown in Table \ref{tab:data-characteristics}, there are only 40 \crisis posts in the training set which is not enough for a supervised learning model to accurately estimate the optimal parameters.

Overall, the results show that both our single and ensemble models can effectively identify posts with critical content (\flagged) with F1 and accuracy of 92\% and 93\%, respectively, on the test set, providing large improvements over the state of the art.

In the rest of this section, we first analyze the effect of different features that we proposed to use for triaging the content severity. Then we analyze the types of errors that our model makes to better understand the robustness of our proposed approach. Finally, using the proposed triaging model, we investigate the potential effect of the mental health forum on the individuals.

\subsection{Feature Analysis}

In the ``Severity Triaging'' Section, we presented our proposed features for the task of triaging the content severity. Table \ref{tab:feat-analysis} shows the effect of each of the features when added to the classification model. We do not show the combinations of features that perform significantly worse than the body of the text. As illustrated, we observe that most of the proposed features have a positive effect on the performance of the system with the exception of skip thought vectors. The bag of words features of the body of the text achieve F1 score of 34.8\% on the test set. Adding contextual features (prior posts by other users and user's previous posts in the thread) improves the results to 38.5\%. Similarly, we observe that addition of forum metadata features (length, kudos, and post views), subjectivity and emotion features, and features from the last sentence also improve the performance. Topic modeling yields further boost to the performance of the system which indicates the effectiveness of latent topics inferred from the forum posts using the LDA model. We observe that \liwc features by themselves do not improve the results as much as topic modeling, however when combined with topic modeling features, greatest improvement is achieved (47.2\% F1). This row (indicated by $\ast$) comprises all features in the single model reported in tables \ref{tab:classification-res} and \ref{tab:per-category}.


\begin{table*}[t]
\centering
\setlength{\tabcolsep}{17pt}
\begin{tabular}{c|cccc|} \cline{2-5}
\cline{2-5}
                                      & \multicolumn{4}{l|}{Macro average over non-\green categories} \\ \hline
\multicolumn{1}{|l|}{Features}        & Acc           & F1            & P             & R             \\ \hline
\multicolumn{1}{|l|}{baseline (body)}            & 87.6          & 34.8          & 33.5          & 36.6          \\
\multicolumn{1}{|l|}{skip thought}    & 87.5          & 33.5          & 33.4          & 34.1          \\
\multicolumn{1}{|l|}{body+contextual} & 90.3          & 38.5          & 36.5          & 40.8          \\
\multicolumn{1}{|l|}{+meta+subj}      & 90.5          & 38.8          & 36.5          & 41.6          \\
\multicolumn{1}{|l|}{+lexical clues}  & 90.9          & 40.2          & 38.3          & 41.3          \\
\multicolumn{1}{|l|}{+last sentence}  & 92.3          & 42.8          & 43.0          & 42.8          \\
\multicolumn{1}{|l|}{+emotion}        & 92.7          & 44.1          & 44.6          & 44.0          \\
\multicolumn{1}{|l|}{+topic}          & 92.9          & 45.8          & 45.5          & 46.2          \\
\multicolumn{1}{|l|}{$-$topic+\liwc}  & 91.8          & 41.9          & 41.7          & 42.6          \\
\multicolumn{1}{|l|}{+topic ($\ast$)}          & 93.9          & 47.2          & 48.9          & 45.8          \\ \hhline{|=|====|}
\multicolumn{1}{|l|}{Ensemble model}  & 94.7          & 50.5          & 51.6          & 49.5          \\ \hline

\end{tabular}
\caption{Effect of each set of features on triaging based on the test set. Numbers show percentages of macro averaged results for the \flagged categories (\crisis $\cup$ \red $\cup$ \amber). Acc: Accuracy, F1: F1-score, P: Precision, R: Recall. Body is the textual body of the post; ``skip thought'' is dense representation of text using skip thought vectors, ``meta'': forum metadata features; ``subj'': subjectivity features; ``topic'': Topic modeling features extracted using LDA, ``\liwc'': Linguistic Inquiry and Word Count features. Plus (+) signs show that the feature is added to the features in the above row and minus ($-$) signs show that the feature is eliminated from the above row. The row shown with ($\ast$) indicates the features (listed in Table \ref{tab:features}) used in the single model in tables \ref{tab:classification-res} and \ref{tab:per-category}. Accordingly, the last row is the ensemble model.}
\label{tab:feat-analysis}
\end{table*}


We build an ensemble of distinct models each of which trained on a different feature set. We experimented with various ensembles of the features. Last row of Table \ref{tab:feat-analysis} shows the performance of the best ensemble model. We do not report other ensembles that resulted in suboptimal performance. The ensemble model that obtains the best results is comprised of 6 different feature sets outlined in Table \ref{tab:features}. As evidenced by Table \ref{tab:feat-analysis}, each of these sets are helpful features that can capture different characteristics of the associated forum post; therefore when combined by ensembling, the weaknesses of single set of features on some instances are compensated by the others. Therefore, as the results show, the ensemble model is more effective in comparison with the single models.

We note that skip thought vectors (second row in Table \ref{tab:feat-analysis}) did not improve the baseline. We also experimented with encoding the prior posts and authors posts with skip thought vectors but we did not observe any improvements. As shown by \citet{kiros2015skip}, when trained on a sufficiently large data, skip thought vectors encode text in dense vectors that can capture underlying semantic and syntactic properties of the text; and thus useful to be used as features for classification. However, in this task we observe that classification using skip thought vectors does not result in any improvements. The lack of improvement by these vectors indicates that the vectors are not able to capture any information beyond what is provided by other features. This could be due to averaging the sentence vectors. We represent the post which consists of several sentences by averaging the vectors corresponding to each sentence; some of the information of the individual sentences might be lost when averaged with other sentences. Therefore, a better approach for composing the post vectors of its constituent sentence vectors could lead to better results.

\subsection{Error Analysis}

Error analysis shows that misclassification of content severity is mainly due to the following reasons:

\begin{enumerate}[labelwidth=!,labelindent=0pt,wide]
\item Brevity of the posts and lack of sufficient background context.

Some \urgent categories that were misclassified are associated with a rather short post from which limited information can be obtained. For example, the following post is taken from a long discussion thread and is labeled as \green by the classifier while the actual label is \red.

\textit{
``I got the reply from x about my complaint. All they did was make excuses for themselves. no help at all.''}

This post on its own does not show any risk to the user. However, reading the entire associated thread in the forum reveals that the author of the post had experienced a problem with their counseling service for their mental distress, and they were in need for mental help and support. To infer this context about this specific post, the immediate surrounding posts are not sufficient and one needs to read the entire conversation.

In the model, we already consider the immediate surrounding posts as the context for the post. However, this may not capture the context in very long discussion threads (such as the above example). When we increased the number of previous posts to be considered as the context, we observed an overall suboptimal performance. This is because, generally, in long threads the discussion tends to change after a few posts. Thus, considering longer window of posts in a thread as context for a target post might result in adding posts that are not necessarily relevant to the target post and consequently introduce noise to the model.

\item Variations in tone.

In some misclassification cases, we observe sudden changes and variations in the tone of the post expressed by the user and that makes it difficult for the learning algorithm to correctly classify the associated severity. For instance consider the following post:

\textit{
`` I went to my favorite show last week and it was amazing. I usually feel very low, specially at nights. This was one of the rare times that I was actually happy for some time... Five days ago at school one classmate of mine bullies me and he shouts that he wishes me dead. I ignored him completely at the moment and I was totally fine. But when I got back home I felt like a total loser and the bad thoughts about myself started coming back.''
}

In this post we observe that the user starts with a positive tone and then it changes to negative. Then the tone switches between positive and negative multiple times. This specific example is an \amber case and the classifier mislabeled it as \red.

\begin{figure*}
\centering
\includegraphics[width=0.4\textwidth]{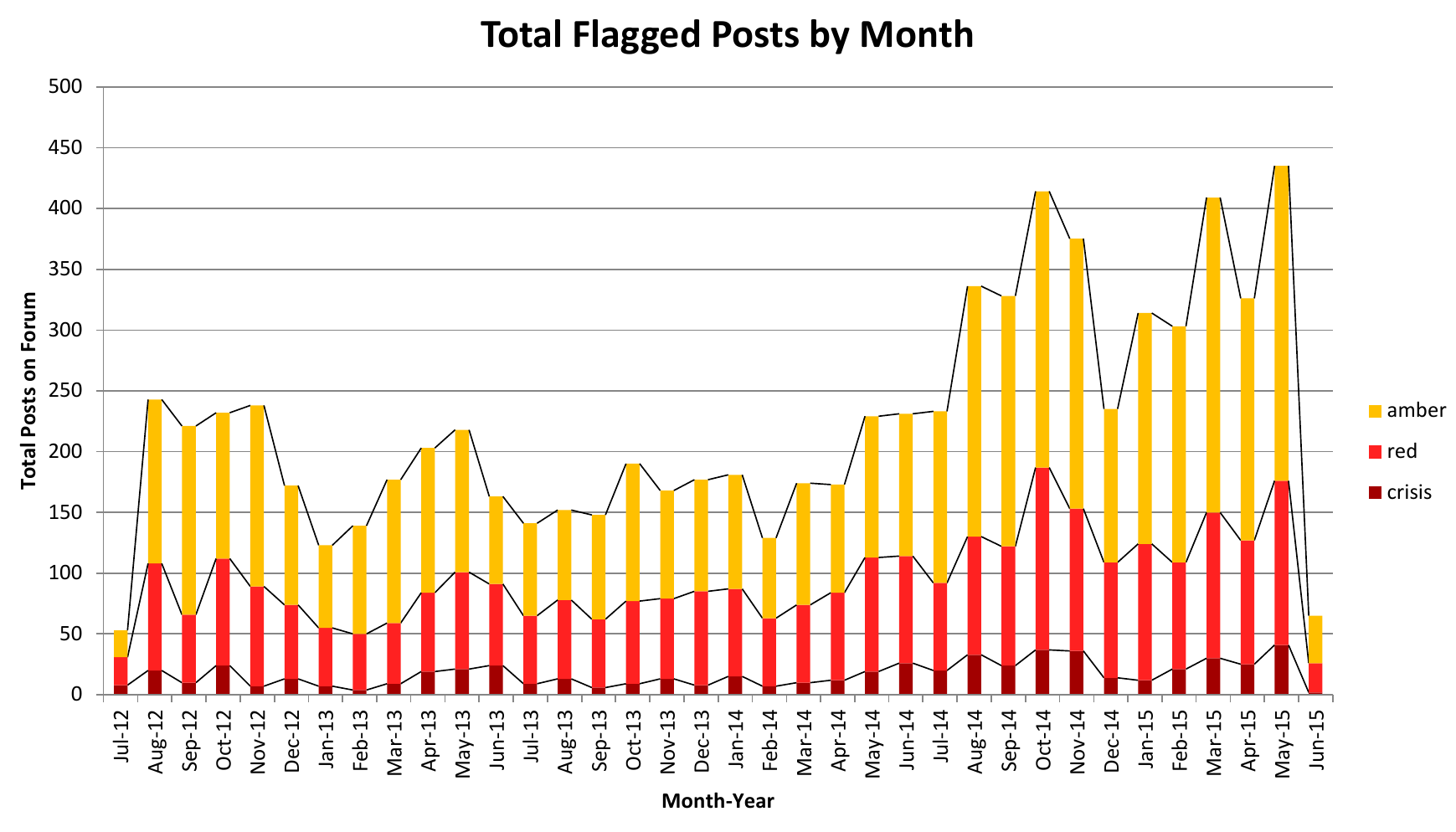}
\caption{Volume of flagged posts on the forum from the time period contained in the dataset}
\label{fig:flagged_posts}
\end{figure*}



\begin{table*}[t]
\centering
\begin{subtable}{.5\linewidth}
\centering
    \begin{tabular}{c|c|c|c|}
    \cline{2-4}
                                     & \multicolumn{3}{c|}{First post} \\ \hline
    \multicolumn{1}{|l|}{Last post} & \flagged    & \green   & Total   \\ \hline
    \multicolumn{1}{|l|}{\flagged}   & 93          & 37       & 127     \\ \hline
    \multicolumn{1}{|l|}{\green}     & 105         & 220      & 325     \\ \hline
    \multicolumn{1}{|l|}{Total}      & 198         & 254      &         \\ \hline
    \end{tabular}
    \caption{\flagged }
    \label{tab:flagged-users-cm}
\end{subtable}%
\begin{subtable}{.5\linewidth}
    \begin{tabular}{c|c|c|c|}
    \cline{2-4}
                                      & \multicolumn{3}{c|}{First Post} \\ \hline
    \multicolumn{1}{|l|}{Last Post}   & \urgent  & non-\urgent  & Total \\ \hline
    \multicolumn{1}{|l|}{\urgent}     & 30       & 16           & 46    \\
    \multicolumn{1}{|l|}{non-\urgent} & 126      & 280          & 406   \\ \hline
    \multicolumn{1}{|l|}{Total}       & 156      & 296          &       \\ \hline
    \end{tabular}
    \caption{\urgent}
    \label{tab:urgent-users-cm}
\end{subtable}
\caption{The number of users by the \flagged \subref{tab:flagged-users-cm} or \urgent \subref{tab:urgent-users-cm} post severity of their first post and their last post. Numbers in cells show the number of users whose first and last post severity corresponds to the associated column and row, respectively. For example 105 in table \flagged \subref{tab:flagged-users-cm} corresponds to the number of users whose first post was \flagged and last post was \green.}
\label{tab:users-cm}
\end{table*}



\begin{table*}[t]
\centering
\begin{subtable}{.45\linewidth}
    \begin{tabular}{c|c|c|c|}
    \cline{2-4}
                                     & \multicolumn{3}{c|}{First month} \\ \hline
    \multicolumn{1}{|l|}{Last month} & \flagged   & \green    & Total    \\ \hline
    \multicolumn{1}{|l|}{\flagged}   & 120        & 46       & 166      \\ \hline
    \multicolumn{1}{|l|}{\green}     & 78         & 208      & 286      \\ \hline
    \multicolumn{1}{|l|}{Total}      & 198        & 254      &          \\ \hline
    \end{tabular}
    \caption{\flagged}
    \label{tab:flagged-users-cm-month}
    \end{subtable}
\begin{subtable}{.45\linewidth}
    \begin{tabular}{c|c|c|c|}
    \cline{2-4}
                                      & \multicolumn{3}{c|}{First month} \\ \hline
    \multicolumn{1}{|l|}{Last month}  & \urgent        & non-\urgent   & Total    \\ \hline
    \multicolumn{1}{|l|}{\urgent}     & 40            & 31            & 71      \\ \hline
    \multicolumn{1}{|l|}{non-\urgent} & 64             & 317           & 381      \\ \hline
    \multicolumn{1}{|l|}{Total}       & 104           & 348           &          \\ \hline
    \end{tabular}
    \caption{\urgent}
    \label{tab:urgent-users-cm-month}
\end{subtable}
\caption{The number of users by average severity of posts in their first and last month of activity in the forum. Numbers in cells show the number of users with average post severity in the first and last month corresponding to the associated column and row. For example 46 in Table (a) corresponds to the number of users whose average post severity in first month was \green and last month was \flagged.}
\label{tab:users-cm-month}
\end{table*}


In the proposed triaging model, we capture the user's final state of the mind by considering features from the last sentence. However, when there are too many tone variations in the post, the exact severity of the post might be misclassified. We note that the size of the training dataset was limited and therefore capturing these subtle cases requires more of similar training instances. Future work could investigate whether these variability of various psychological variables (e.g. tone) can be considered as a risk factor for individuals.

\item Long posts with only a small part containing concerning content.

In a few long posts, we observe only a small part showing signs of distress to the user, while the rest of the post has a neutral to positive tone. A misclassified example with actual label of \red is shown below (Parts indicated with [...] are omitted for brevity):

\setul{0.5ex}{0.1ex}
\definecolor{orange}{rgb}{0.55,0.55,0.55}
\setulcolor{orange}

\textit{
 ``This book series is a roller coaster. Maze runner series, I'm onto the prequel book now. They are amazing [...] I've always been too resilient. \ul{I just hate everything and it confuses me. Maybe I'm tired of all this and want to do something.. I just... nothing is set}. Yesterday Lora called and we talked like a lot about school, friends [...] It feels good to say, or type, all this.''
\textit}

This snippet is from a much longer post and as it can be observed, only the underlined part contains content that indicate mental distress to the user.

In such posts, the effect of the small negative part of the post is played down by the larger dominant neutral tone and therefore the model could mispredict this. In this case although still correctly identified as critical, the classifier misclassifies the severity level as \amber instead of \red.

\end{enumerate}

Overall, most of classification errors occur within the \flagged category; there are very few cases in the \flagged posts that are missed by the classifier and labeled as \green. This can also be observed in Table \ref{tab:classification-res} in \flagged category performance which obtains F1 and accuracy scores of 92.2\% and 93.4\%, respectively. Our results are encouraging since they show that the model can effectively capture \flagged posts, i.e. all posts that indicate some signs of harm to the user.


\subsection{User Analysis}

We study the user content severity in the forum over time to analyze if it is helpful to the individuals. For the purposes of user analysis, we mostly rely on the binary classification of \urgent (\crisis and \red) vs. non-\urgent, and \flagged (\crisis $\cup$ \red $\cup$ \amber) vs. \green categories. In these categories, as shown in Tables \ref{tab:classification-res} (a and b), the ensemble classification model obtains F-1 scores of 90\% and 75\% respectively (accuracy of 91\% and 93\%) and thus it is relatively reliable for studying larger scale trends of content severity in the entire forum. Figure \ref{fig:flagged_posts} shows the results of severity triaging throughout all the posts in the dataset. As illustrated, there is a steady increase in the amount of \flagged posts. Given this trend, we examine patterns of post severity to understand the effects that the forum might have on the individuals. Specifically, we investigate the following research questions:

\begin{enumerate}[wide, labelwidth=!,labelindent=0pt, label=\textbf{Q-{\arabic*}}.]
\item \textbf{Does engaging with the forum have a positive effect on the users?}

Our analysis indicated a decline in the average content severity over time, which may indicate a positive effect of the forum on its users. This correlative effect suggests that further controlled trials should be conducted to carefully ascertain the causal nature of this relationship.


The dataset includes posts from the forum in a time window of 36 months during which we quantify the behavior of users. To measure the relation of user interaction with the forum, we split the users into two groups. Users are considered active if they have posted for two or more months on the forum, and inactive if they had only posted during a single month. We only consider active users for the analysis because for inactive users, the activity period of one month is too short to present a significant relation. In these 36 months, there are a total of 452 active users and 1,195 inactive users. We analyze the severity of the first post and last posts of users, average post severity during their first and last months of activity and finally, the trend lines of severity during entire time of interaction with the forum.

Tables \ref{tab:flagged-users-cm} and \ref{tab:urgent-users-cm} show the number by the severity of their first and last posts on the forum. A Chi-square test on the contingency tables was performed to ensure that the difference between the cells are interpretable. For both table \ref{tab:flagged-users-cm} and \ref{tab:urgent-users-cm} we found significant interaction, $\chi^2=58.4$, $p<.001$ and $\chi^2=21.4$, $p<.001$, respectively.
In general, we observe that the users' last posts tend to be of lower severity than their first post. 81\% of users whose first post received an \urgent label had a final post with a non-\urgent label. Only 10\% of users whose first post was non-\urgent had a final post of \urgent. In both the \flagged and \urgent matrices, there were more users whose final posts was \green or non-\urgent than users who had \flagged or \urgent first posts.

Tables \ref{tab:flagged-users-cm-month} and \ref{tab:urgent-users-cm-month} show the comparison of the average user content severity in the first and last month of users' activity in the forum (Chi-square test showed that the results are interpretable with a significant difference of $\chi^2$=86.47 ($p<0.001$) and $\chi^2$=52.82 ($p<0.001$) for Tables \ref{tab:flagged-users-cm-month} and \ref{tab:urgent-users-cm-month}, respectively.). We observe a similar positive trend in the \urgent category; in the \flagged category, the number of users whose average initial content and last content is \flagged (120 users) is more than those whose content is shifted from \flagged to \green (78 users). However, there are very few \green users whose content eventually turned \flagged (46 users). Furthermore, the total number of users with first month \flagged posts (198) is higher than number of users with last month \flagged posts (166).  These results also indicate that users' last posts tend less severe than their first posts.


\begin{figure}[]
\centering
\includegraphics[width=0.4\textwidth]{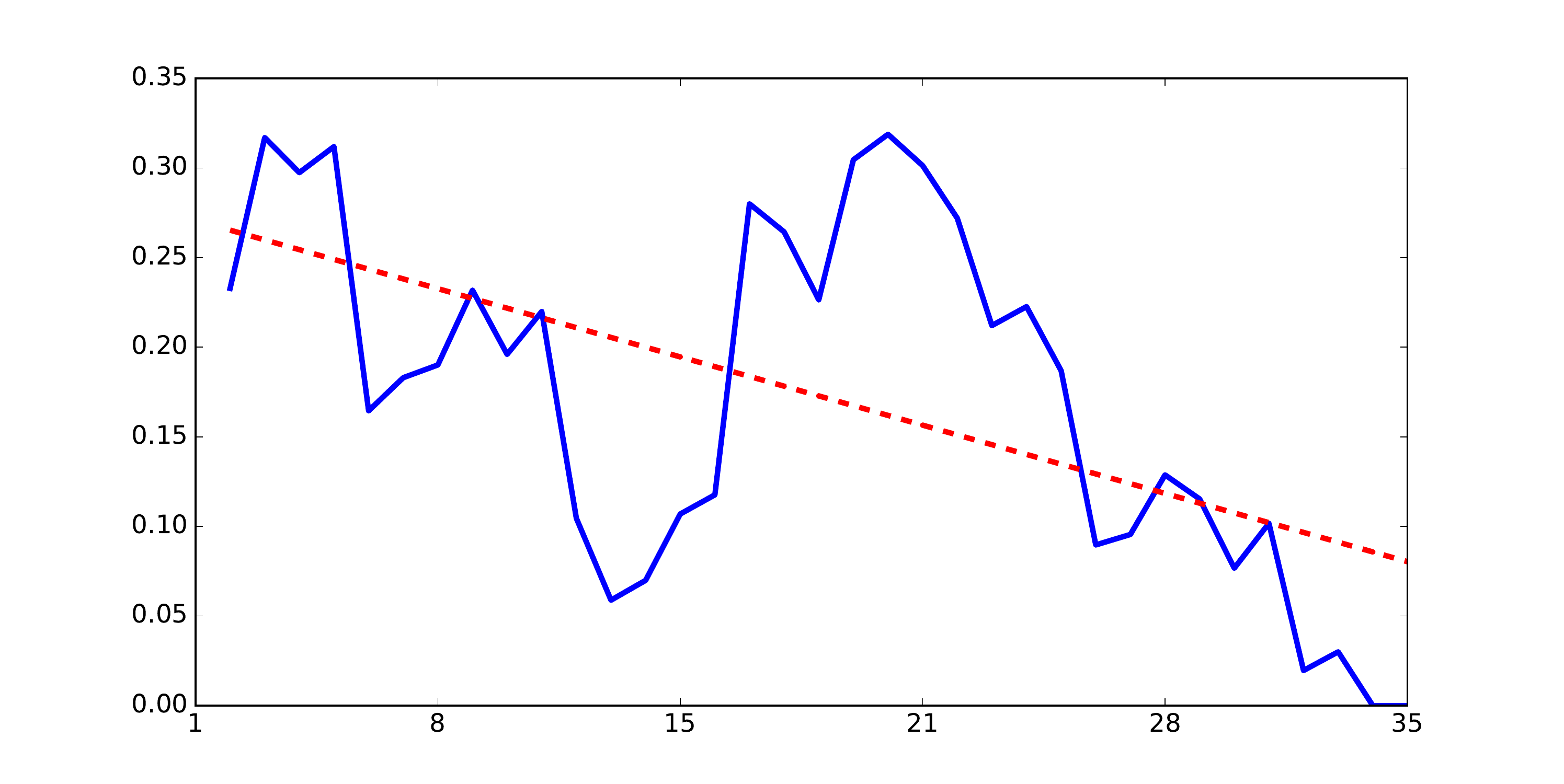}
\caption{Example trend line of user post severity over time. $x$ axis shows the month of activity. $y$ is the average content severity in the month.}
\label{fig:trend-exmpl}
\end{figure}

\begin{table}[t]
\centering
\begin{tabular}{|l|c|c|}
\hline
               & Avg.  & Std dev. \\ \hline
Positive trend & 0.68  & 0.34     \\ \hline
Negative trend & -0.72 & 0.32     \\ \hline
\end{tabular}
\caption{The average (Avg.) and standard deviation (Std dev.) of the r values of the trend lines for the positive and negative trends.}
\label{tab:goodness-of-fit}
\end{table}

\begin{table*}[t]
\centering
\setlength{\tabcolsep}{2pt}
\begin{tabular}{|c|c|c|c|c|c|c|c|c|}
\hline
          & \multicolumn{4}{c|}{\flagged vs \green}     & \multicolumn{4}{c|}{Fine-grained severity}    \\ \hline
Threshold & Avg.   & Stdev. & \# positive & \# negative & Avg.   & Stdev. & \# positive & \# negative \\ \hline
0.02      & -0.096 & 0.370  & 90          & 113         & -0.068 & 0.236  & 76          & 120         \\ \hline
0.05      & -0.134 & 0.430  & 60          & 86          & -0.104 & 0.285  & 43          & 84          \\ \hline
0.10      & -0.177 & 0.458  & 41          & 72          & -0.129 & 0.320  & 32          & 65          \\ \hline
0.15      & -0.224 & 0.510  & 30          & 57          & -0.159 & 0.350  & 23          & 53          \\ \hline
None      & -0.044 & 0.221  & 167         & 272         & -0.032 & 0.151  & 153         & 298         \\ \hline
\end{tabular}
\caption{Analysis of trend lines of severity over time for active users. \flagged vs \green indicates the trend change between \flagged and \green categories while fine-grained severity is for all 4 severity categories. \textit{Avg.} shows the average of the slope of the trendlines. \textit{Stdev.} is the standard deviation of the slope of the trendlines. \textit{\#positive} shows the number of users with positive slope of trendline. \textit{\#negative} shows the number of users with negative slope of trendline. Negative (positive) slope of trend line shows decreased (increased) content severity of the user over time. Threshold is used to filter out the effect of the flat trendlines; the considered trend lines in each row have an absolute value of slope greater than the value of the threshold in that row. Overall, the Table indicates that the content severity for majority of the users with non-flat trend line has decreased over time.}
\label{tab:trendlines}
\end{table*}

We believe this is because many users join this type of forums to get immediate support for a moment of crisis or acute mental distress. After some time, this initial distress is decreased, as reflected in the patterns of post severity. That is why the initial activity of users in general tend to be more severe than their final posts.  We believe this could be for the following reasons: \textit{(i)} Pattern of post severity drops off once the user is in a more stable mental state compared with their initial state of crisis. \textit{(ii)} Interaction with the forum and engaging in discussion with other forum users might have resulted in reducing the acute distress in users (verifying the exact causal relation requires further user level controlled trials).

In addition to first and last months of activity, we also analyze the trends throughout the entire time of user activity. To do so, we consider the average severity of the posts in each month as a data point for that month, and we then fit a trend line to the data points. We consider the following numeric values for each category to be able to quantify the average severity in each month: \crisis = 1.0, \red = 0.66, \amber = 0.33, \green = 0.0.
Using these numeric equivalent of severity classes, for each user, we associate an average severity for all their posts in each month. Then, we fit a linear model on this data to show the trend line of the content severity over time.
Figure \ref{fig:trend-exmpl} shows a sample plot of the post severity for a user over time and its associated trend line.

To fit an appropriate trend line to the data, we minimize the squared error between the target trend line and the actual severity data points. Specifically, the equation of a trend line for variable $x$ is given by $p(x) = m.x + b$ where $m$ and $b$ are the slope and intercept of the line, respectively. A negative (positive) trend line slope indicates that overall, the severity of user content has declined (increased). Given $D$ severity data points $\{(x_i,y_i)\}_{i=1}^D$, the values of $m$ and $b$ are found by minimizing the squared error over the data:
\begin{equation}
E=\sum\limits_{i=0}^D|p(x_i)-y_i|^2
\end{equation}


\begin{table*}[t]
\centering
    \begin{subtable}{.5\linewidth}
      \centering
        \begin{tabular}{c|c|c|}
        \cline{2-3}
                                        & \multicolumn{2}{c|}{First Month} \\ \hline
        \multicolumn{1}{|l|}{Last Month} & \flagged         & \green         \\ \hline
        \multicolumn{1}{|l|}{\flagged}   & 3.47            & 5.15          \\
        \multicolumn{1}{|l|}{\green}     & 3.62            & 7.02          \\ \hline
        \end{tabular}
        \caption{\flagged}
        \label{tab:flagged-monthly}
    \end{subtable}%
    \begin{subtable}{.5\linewidth}
      \centering
        \begin{tabular}{c|c|c|}
        \cline{2-3}
                                         & \multicolumn{2}{c|}{First Month} \\ \hline
        \multicolumn{1}{|l|}{Last Month}  & \urgent       & Non-\urgent       \\ \hline
        \multicolumn{1}{|l|}{\urgent}     & 3.14          & 5.11             \\ \hline
        \multicolumn{1}{|l|}{Non-\urgent} & 3.28         & 6.48             \\ \hline
        \end{tabular}
        \caption{\urgent}
        \label{tab:urgent-monthly}
    \end{subtable}
    \caption{The average number of months the users stayed active in the forum based on the average severity of their content in the first and last months of activity.}
\end{table*}



\begin{table*}[t]
\centering
\begin{tabular}{ |c|c|c|c|c|c|  }
 \hline
 \multicolumn{6}{|c|}{Moderator Response Time} \\
 \hline
  & Total & Number & Percentage & Average Time & Stdev Time \\ \hline
  \crisis & 608 & 147 &24.18\% &4.21 & 5.71 \\
\red & 2798 & 931 & 33.27\% & 4.53 & 6.17 \\
\amber & 4642 & 1435 & 28.05\% & 4.46 & 6.60 \\
\green & 57707 & 892 & 1.55\% & 3.76 & 6.07 \\ \hline
\urgent & 3406 & 1078 & 37.96\% & 4.37 & 5.94 \\
\flagged & 8048 & 2513 & 37.88\% & 4.40 &6.16 \\
 \hline
\end{tabular}
\caption{Time in hours. Average response time when a moderator was the first to respond.}
\label{tab:mod-resp-time}
\end{table*}


To check if a linear model is a applicable for our case, we calculated the r values associated with the trendlines. In particular, for each user we calculated the r value of their content severity trend lines and we calculated the average and the standard deviation of these values (Table \ref{tab:goodness-of-fit}). As illustrated, the average of r values are 0.68 (-0.72) for the positive (negative) trends which is around 0.7 (-0.7). Absolute r values greater than 0.5 indicate high to strong linear relationship in the data \cite{tabachnick2001using}. Thus, linear trend analysis is a reasonable fit to this data.


%

To analyze overall trends in the content severity, we calculate the content severity trend line for each user and then analyze the overall trend line statistics for the users.
We observed that many users have steady trend lines with a slope of near zero. To eliminate the noise caused by these neutral trends from our analysis, we filter out the users whose content severity trend lines are essentially flat. These users are either the moderators of the forum or are users that show consistent behavior over time. We then analyze how the content severity of the other users with varying content severity changes over time. Table \ref{tab:trendlines} shows the statistics for all the trends lines among all the active users.
To eliminate trend lines having a slope near zero, we consider a threshold. We analyze results based on different values of this threshold. For example, for the threshold $\tau$, the corresponding row on the Table only considers trend lines with slope $m$ such that $m<-\tau$ or $m>\tau$ and filters out all other lines having $|m|<=\tau$. We also show the results in the case that there is no threshold (last row of the Table). \flagged vs \green corresponds to plots with numeric severity value of 1.0 for a \flagged post and 0.0 for a \green post; Fine-grained severity categories corresponds to plots with following numerics severity values: \crisis = 1.0, \red = 0.66, \amber = 0.33, \green = 0.0. As illustrated in Table \ref{tab:trendlines}, we observe an average negative trend line slope for all the values of the threshold. This indicates a decline of average content severity among all the users. Furthermore, we observe that majority of users have a trend line with a negative slope and thus, decreasing severity of content.

These results indicate that overall there is a decline in the content severity of the users as they interact with the forum, which could be due to the potential positive effect of the forum on its users. This effect could be attributed to the users expressing their feelings and emotions, receiving support and feedback from the moderators, and discussing issues with users experiencing similar problems. However, we note that here we only observe the negative trend of content severity; to study the exact causal relationship between interaction with the forum and content severity, further controlled trials on the forum users should be conducted.


\item \textbf{What effect does the duration of engagement with the forum have on the users?}

We analyze how the duration of a user's engagement with the forum impacts the severity of their posts over time. Tables \ref{tab:flagged-monthly} and \ref{tab:urgent-monthly} show that users with a first month severity of \flagged or \urgent posts interacted with the forum for 3-4 months, while other users interacted with the forum for 5-7 months. These tables are essentially showing that users with less critical posts in the first month tend to interact with the forum in a more long-term basis in comparison with users whose initial posts are critical. The difference in the duration of user interaction by their initial content severity indicates that there are users who visit the forum for immediate assistance in a critical moment and those who use the forum as a longer-term support resource.  This result suggests that users whose first posts are more severe could be on the forum for immediate support and will only stay active until their critical mental state reaches a safe equilibrium again. In contrast, the users whose first month is \green or non-\urgent may be seeking a long-term resource and a community of users with shared experiences.

This difference between the activity period of users by their initial content reveals an opportunity for moderators to improve their response time to \flagged and \urgent posts.
Faster moderator attention to \flagged and \urgent posts would provide better quality of help to these short-term users and encourage them to further interact with the forum for receiving support. Triaging the forum posts to allow moderators improve their response time would benefit all user groups, and particularly users who currently visit the forum for an immediate support.

\item \textbf{What is the effect of moderator response time on the user's forum behavior?}

Since the focus of this research is on triaging the severity of mental health forum posts, we seek to understand how quickly moderators are currently responding to posts by their severity. Table \ref{tab:mod-resp-time} shows the average time for a moderator to respond, as well as the percentage of cases in which the moderators were the first to respond to a user. It shows that in cases where a moderator was the first to respond to a \flagged or \urgent post, they took on average more than four hours to respond. Unfortunately, four hours might be too long for users with imminent risks and it is very important to reduce this response time to prevent a potential self harm. Additionally, we observe that moderators are the first responders on less than 33\% of non-\green posts, meaning the other forum users responded to majority of posts earlier than moderators. This further stresses the value of triaging content severity, so that moderators can quickly respond to critical posts rather than having to identify such posts on the forum manually.


\end{enumerate}


\section{Conclusions}

We presented an approach for triaging the content severity in mental peer support forums with a specific goal of identifying cases with potential risk of self-harm. Triaging the content severity helps the forum moderators to locate the critical cases and attend to them as soon as possible. We used a feature-rich classifier with various sets of features including psycholinguistic, contextual, topic modeling and forum metadata features for triaging the content into different severity categories. In addition to a single classifier, we also built an ensemble classifier by using different sets of features. We evaluated our approach on the data from ReachOut.com, a large mental health forum. We showed that our approaches can effectively improve over the state-of-the-art by large margins (up to 17\% macro-average F1 scores of critical categories). We showed that the content severity of the users tend to decrease as they interact with the forum. Results further indicated that there is a need for effective and efficient triaging of forum post data to assist the moderators in attending the users with potential risk of self-harm.

The impact of this research is important from two perspectives. It stresses the importance of mental health forums as a support platform for users with mental health problems. It furthermore provides an efficient and effective way for moderators to asses the content severity of the forum, and consequently help individuals in need and prevent self harm incidents.

\section{Acknowledgments}
This work was partially supported by National Science Foundation (NSF) through grant CNS-1204347.


\bibliographystyle{acl_natbib}
\bibliography{refs}

\end{document}